%% file: blueberry_harvesting.tex
\documentclass[lettersize,journal]{IEEEtran}
\usepackage{amsmath,amsfonts}
\usepackage{algorithmic}
\usepackage{array}
\usepackage[caption=false,font=normalsize,labelfont=sf,textfont=sf]{subfig}
\usepackage{textcomp}
\usepackage{stfloats}
\usepackage{url}
\usepackage{verbatim}
\usepackage{graphicx}
\usepackage{multirow}    
\usepackage{tikz}
\usepackage{booktabs}
\usepackage{graphicx}
\usepackage{afterpage}
\usepackage{siunitx}
\DeclareSIUnit{\pound}{lb}
\DeclareSIUnit{\tonne}{t}
\usepackage{float}
\usepackage[normalem]{ulem}
\usepackage[flushleft]{threeparttable}
\def\BibTeX{{\rm B\kern-.05em{\sc i\kern-.025em b}\kern-.08em
    T\kern-.1667em\lower.7ex\hbox{E}\kern-.125emX}}
\usepackage{balance}
\usepackage[table]{xcolor}
\usepackage{siunitx}
\usepackage{hyperref} 
\definecolor{secrow}{HTML}{DCE9F5}

\begin{document}
\bstctlcite{IEEE:BSTcontrol} 
\title{CLASP: A Cluster-Level Autonomous Selective Picking Robot with a Soft Rolling-Band Gripper for Fresh-Market Blueberry Harvesting}

\author{Yixuan Xia\IEEEauthorrefmark{1}, Yilin Cai\IEEEauthorrefmark{1}, Natalia Belen Espinoza, Changying Li, Zilfina Rubio Ames, Xin Zhang, Yue Chen

\thanks{This work was supported by Georgia Tech IRIM seed grant, and partially by the USDA-NIFA’s National Robotics Initiative Award 2022-11065 in collaboration with the National Science Foundation (Corresponding
author: Yue Chen). Yixuan Xia and Yilin Cai contributed equally to this work. 

Yixuan Xia and Yilin Cai are with the George W. Woodruff School of Mechanical
Engineering, Georgia Institute of Technology, Atlanta, GA 30332 USA (e-mail: yxia339@gatech.edu; yilincai@gatech.edu).


Changying Li is with the Bio-Sensing, Automation, and Intelligence Laboratory, Department of Agricultural and Biological Engineering, Institute of Food and Agricultural Sciences, University of Florida, Gainesville, Florida 32611, USA (e-mail: cli2@ufl.edu).

Natalia Belen Espinoza and Zilfina Rubio Ames are with the Department of Horticulture, University of Georgia, Tifton–CAES Campus, 2360 Rainwater Road, Tifton, GA 31793, USA (e-mail: nbe77578@uga.edu; zilfina.rubioames@uga.edu). 

Xin Zhang is with the School of Environmental, Civil, Agricultural and Mechanical Engineering, University of Georgia, Athens, GA 30602 USA (e-mail:
xin.Zhang2@uga.edu). 

Yue Chen is with  the Wallace H. Coulter Department of Biomedical Engineering, Georgia Institute of Technology/Emory University, Atlanta, GA 30332 USA (e-mail: yue.chen@bme.gatech.edu). 

}}

\maketitle
\begin{tikzpicture}[remember picture,overlay]
  \node[anchor=north, yshift=-8mm] at (current page.north) {%
    \parbox{00.8\textwidth}{\centering\footnotesize
      This work has been submitted to the IEEE for possible publication.
      Copyright may be transferred without notice, after which this
      version may no longer be accessible.}};
\end{tikzpicture}
\begin{abstract}
Fresh-market blueberries require selective, gentle picking, which is labor-intensive and expensive. Over-the-row machine harvesters are fast but non-selective, bruising mixed-ripeness fruit and limiting yield to the processing market. Selective robotic harvesters typically target individual fruits rather than fruit clusters, which limits harvesting efficiency for small, densely clustered blueberries. This paper presents CLASP, a Cluster-Level Autonomous Selective Picking robot with a Soft Active Rolling-Band Gripper (SARB-Gripper) for fresh-market blueberry harvesting. Two compliant bands envelop the cluster and roll against berries, detaching mature berries in sequence, while closed-loop regulation of the pulling force keeps the applied load below the immature detachment threshold. A global-to-local perception pipeline pairs an eye-to-hand camera for global cluster detection and target selection with an eye-in-hand camera for localization and cluster orientation estimation. Field measurements confirm a clear detachment-force separation between mature and immature fruit. The SARB-Gripper achieves repeatable pulling-force regulation across varying object sizes, with a coefficient of variation below \SI{3.7}\%, supporting selective cluster-level harvesting.
In end-to-end field trials, CLASP autonomously grasped 23 of 25 presented clusters (\SI{92}{\percent}). At a component cost of ~\$3,326 per unit, CLASP offers a scalable approach to selective cluster-level harvesting for fresh-market blueberries.
Project website: \protect\href{https://stophere1.github.io/CLASP_page/}{https://stophere1.github.io/CLASP\_page/}.
\end{abstract}


\input{introduction}
\input{relatedworks}
\input{design}
\input{perception}

\input{control}
\input{analysis}
\input{experimentsandresults}

\input{discussion}
\section*{Acknowledgments}
\noindent The authors thank Dr.~Guihong Bi and Dr.~Eric Stafne of Mississippi State University for providing farm access for imagery collection, and Thevathayarajh~Thayananthan and Jonathan~Harjono of the University of Georgia for assistance with field data collection.

\bibliography{citations}
\bibliographystyle{IEEEtran}
\vspace{-33pt}
\begin{IEEEbiography}[{\includegraphics[width=1in,height=1.25in,clip,keepaspectratio]{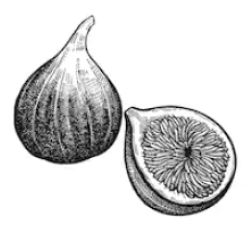}}]{Yixuan Xia} received the B.S. degree in Robotics from Southern University of Science and Technology, Shenzhen, China, and the M.S. degree in Mechanical Engineering from Carnegie Mellon University. He is currently working toward the Ph.D. degree in Robotics with the Georgia Institute of Technology, Atlanta, GA, USA.

His research interests include perception and control fosr mobile manipulation.
\end{IEEEbiography}
\vspace{-33pt}
\begin{IEEEbiography}[{\includegraphics[width=1in,height=1.25in,clip,keepaspectratio]{./figures/bios/fig1}}]{Yilin Cai} received the B.S. degree from Shanghai Jiaotong University, Shanghai, China, and the M.S. degree in Robotics System Development from Carnegie Mellon University. He is currently working toward the Ph.D. degree in Robotics with the Georgia Institute of Technology, Atlanta, GA, USA.
His research interests include soft robotics and robot manipulation.
\end{IEEEbiography}
\vspace{-33pt}
\begin{IEEEbiography}[{\includegraphics[width=1in,height=1.25in,clip,keepaspectratio]{./figures/bios/fig1}}]{Natalia Belen Espinoza} is a Research Assistant at the University of Georgia.  Natalia completed her Master’s degree in Crop and Soil Sciences with an emphasis in Sustainable Agriculture from the University of Georgia. She did her undergraduate studies in Ecuador in Agroindustrial Engineering.

She is working on improving the fertilization regime and soil amendments for blueberries under the guidance of Dr. Zilfina Rubio Ames.

\end{IEEEbiography}
\vspace{-33pt}
\begin{IEEEbiography}[{\includegraphics[width=1in,height=1.25in,clip,keepaspectratio]{./figures/bios/fig1}}]{Changying Li} received the Ph.D. degree in Agricultural and Biological Engineering from The Pennsylvania State University, University Park, PA, USA, in 2006. He is currently a Professor with the Department of Agricultural and Biological Engineering, University of Florida, Gainesville, FL, USA.

His research interests include agricultural robotics, machine learning and computer vision, and phenomics.

\end{IEEEbiography}
\vspace{-33pt}
\begin{IEEEbiography}[{\includegraphics[width=1in,height=1.25in,clip,keepaspectratio]{./figures/bios/fig1}}]{Zilfina Rubio Ames} received the M.S. and Ph.D. degrees in Horticultural Science from the University of Florida, Gainesville, FL, USA. She is currently an Assistant Professor and Small Fruit Extension Specialist with the Department of Horticulture, University of Georgia, Tifton, GA, USA.

Her research interests include whole-plant physiology, nutrient management, and cultural practices for blueberries and other small fruits.

\end{IEEEbiography}
\vspace{-33pt}
\begin{IEEEbiography}[{\includegraphics[width=1in,height=1.25in,clip,keepaspectratio]{./figures/bios/fig1}}]{Xin Zhang} received the Ph.D. degree in Biological and Agricultural Engineering from Washington State University, Pullman, WA, USA. She is currently an Assistant Professor with the School of Environmental, Civil, Agricultural and Mechanical Engineering, University of Georgia, Athens, GA, USA.

Her research interests include agricultural automation and robotics.
\end{IEEEbiography}
\vspace{-33pt}
\begin{IEEEbiography}[{\includegraphics[width=1in,height=1.25in,clip,keepaspectratio]{./figures/bios/fig1}}]{Yue Chen} received Ph.D. degree from Vanderbilt University, Nashville, TN, USA. He is currently an Associate Professor with the Wallace H. Coulter Department of Biomedical Engineering, Georgia Institute of Technology and Emory University, Atlanta, GA, USA.

His research interests include the design, fabrication, modeling, and control of medical and soft robotic systems.

\end{IEEEbiography}
\vfill
\end{document}

%% file: introduction.tex
\section{Introduction}
\label{sec:introduction}

U.S. cultivated blueberry production reached about 358,000 tons with a farm-gate value of roughly \$1.15 billion in 2024~\cite{usdaers2025blueberry}. Yet harvesting for the fresh market remains almost entirely manual~\cite{intelligentrobotsreview}, because within a single cluster blueberries ripen asynchronously. Pickers must pass through the field repeatedly, harvesting only ripe fruit while leaving the rest to mature. Harvesting is thus the most labor-intensive and costly stage of fresh-market production, and as seasonal farm labor grows scarce, the pressure to automate it intensifies. 

Despite this strong need for automation, fresh-market blueberry harvesting presents three coupled challenges.
The first is \emph{gentle handling}: conventional over-the-row harvesters shake blueberries from the bush, but studies show that mechanically harvested berries exhibit \SIrange{16}{26}{\percent} severe bruising even with soft-catch surfaces, compared with only \SIrange{1}{4}{\percent} for hand-harvested berries~\cite{mechanicalharvester}. This damage confines shaking largely to the processing market and leaves its fresh-market economics marginal~\cite{mechanicalharvesterrevenue}. The second is \emph{ripeness selectivity}: ripe and unripe berries are intermingled on a common stem, so a harvester must remove mature fruit while leaving immature fruit attached for later passes. 
The third is \emph{dense cluster geometry}: blueberries are small and grow in tight clusters, where adjacent fruit contact and occlude one another. This structure makes it difficult to perceive and manipulate individual berries independently, limiting the applicability of per-fruit harvesting strategies developed for larger, sparsely distributed fruits. Blueberry harvesting must therefore operate at cluster level, while selectively detaching mature fruit and minimizing mechanical loading on the retained and harvested berries.

These three requirements are coupled, yet existing autonomous harvesting systems do not address them simultaneously. 
Most robotic harvesting systems developed for larger, sparsely distributed fruits follow a per-fruit paradigm, in which each target is individually localized and manipulated by the robot~\cite{appleharvester, li_multi-arm_2023, zhu_advancement_2025}. Meanwhile, compliant and soft grippers are developed to handle fruit gently~\cite{softgripper}, but are likewise typically designed to grasp one individually localized fruit at a time.
This approach does not readily extend to blueberries, where harvesting must instead operate on multiple fruit within a cluster.
Such cluster-level interaction introduces an additional requirement for selectivity, in which the applied load must be sufficient to detach mature berries while allowing immature berries to remain attached.

The key challenge is therefore to move beyond individual-fruit manipulation toward a cluster-level operating principle that achieves selective detachment through controlled force rather than precise berry-level localization and manipulation.

To address these challenges, we present \textbf{CLASP}, a \textbf{C}luster-Level \textbf{A}utonomous \textbf{S}elective \textbf{P}icking robot for fresh-market blueberry harvesting, shown in Fig.~\ref{fig:hardware_design}. Rather than localizing and manipulating each berry individually, CLASP uses a soft active rolling-band gripper (SARB-Gripper) to engage multiple berries simultaneously and progressively detach them through continuous rolling contact. Its pulling force is regulated using servomotor current feedback without a dedicated force/torque sensor. By maintaining the applied force within the detachment-force margin between mature and immature berries, the gripper selectively removes only the mature ones. Autonomous harvesting is enabled by a global-to-local perception and control pipeline. An eye-to-hand camera provides coarse cluster localization, while an eye-in-hand camera refines cluster geometry and assesses harvest completion. A finite-state controller coordinates perception, arm motion, gripper force regulation, and harvesting actions. 

The main contributions of this work are:
\begin{itemize}
  \item a cluster-level harvesting robotic platform combining a compliant rolling-band gripper with sensorless force control for gentle, ripeness-selective detachment;
  \item a multi-season field dataset, together with a global-to-local perception pipeline for cluster detection, localization, orientation estimation, and harvesting progress monitoring;
  \item an integrated autonomous harvesting system validated under field conditions, achieving a \SI{92}{\percent} cluster-level harvesting success rate.
\end{itemize}

%% file: relatedworks.tex
\section{Related Works}
\label{sec:relatedwork}
\begin{table}[!t]
\centering
\caption{Field-evaluated robotic harvesting systems.}
\label{tab:systems}
\footnotesize
\setlength{\tabcolsep}{2pt}
\begin{tabular*}{\columnwidth}{@{\extracolsep{\fill}}llccl@{}}
\hline
Crop & Manipulator\textsuperscript{$\ast$} & Success\ (\%) & Cycle (s) & Setting \\
\hline
Apple~\cite{li_multi-arm_2023}        & Custom multi-arm\textsuperscript{$\dagger$} & 71.3--80.5 & 5.8--6.7 & Orchard \\
Apple~\cite{zhu_advancement_2025}     & Custom 4-DOF$\times$2 & 80.7/79.7 & 5.97 & Orchard \\
Strawberry~\cite{PPParm}              & Custom dual-arm       & 50--97.1\textsuperscript{a} & 4.6--6.1\textsuperscript{b} & Polytunnel \\
Strawberry~\cite{strawberryharvester} & Commercial 7-DOF      & 83\textsuperscript{c} & --- & Glasshouse \\
Tomato~\cite{ansari_integrated_2026}  & Commercial 5-DOF\textsuperscript{$\dagger$} & $\sim$80 & 24.34 & Lab \\
Pumpkin~\cite{roshanianfard_performance_2018} & Custom tractor-mtd & 92 & 35.1--58.7 & Field \\
\textbf{Blueberry (ours)}             & \textbf{Custom 6-DOF} & \textbf{92}\textsuperscript{d} & --- & \textbf{Field} \\
\hline
\end{tabular*}\\[2pt]
\raggedright\scriptsize
\textsuperscript{$\ast$}Tiers inferred from manipulator class where cost is unreported;
\textsuperscript{$\dagger$}Model not stated in abstract.
\textsuperscript{a}First attempt, growth-situation dependent.
\textsuperscript{b}Manipulation only.
\textsuperscript{c}Of fruits judged pluckable.
\textsuperscript{d}Success when the cluster is autonomously grasped.
\vspace{-18pt}
\end{table}
\subsection{Manipulation Hardware for Selective Fruit Harvesting}
Selective harvesting in dense canopies imposes two key requirements on robotic hardware: sufficient manipulator workspace and dexterity at an affordable cost, and compliant interaction between the end-effector and fruit. Most existing fruit-harvesting robots rely on commercial industrial arms, which provide the required dexterity and positioning accuracy. But their high cost limits multi-arm deployment, particularly on small farms~\cite{usdaers2022labor}. Lower-DoF manipulators are more compact and cost-effective but typically sacrifice workspace and orientation dexterity~\cite{PPParm, roshanianfard_performance_2018}, both critical for approaching clusters in occluded canopies while avoiding surrounding branches and foliage.

At the fruit interface, the end-effector must accommodate the delicate and irregular geometry of berry clusters.
Compliant grippers reduce fruit damage and compensate for geometric uncertainty through tendon-driven~\cite{9684953, gunderman2024kinetostatics}, suction-based~\cite{krueger2026analysissensorselectionfruit}, or linkage-based hybrid~\cite{11237059} designs. Soft compliance combined with actively driven contact surfaces offers a promising alternative for fruit handling beyond conventional grasping~\cite{cai2023hand, he2025grasping}.
However, these approaches have not specifically addressed force-regulated, maturity-selective fruit detachment at the cluster level.

\subsection{Perception in Cluttered Agricultural Environments}
For autonomous harvesting, fruit detection alone is insufficient. Perception must also provide spatial information accurate enough to guide manipulation. Vision-based harvesting systems commonly combine learning-based detection and segmentation, such as YOLO-family models, with RGB, stereo, or RGB-D sensing~\cite{owais_agrivision_2025}.
In field environments, however, harvesting performance is often constrained by occlusion, canopy structure, and target accessibility in addition to detector accuracy.
For example, an apple harvesting pipeline reported a decrease in success rate from \SI{82.4}{\percent} in a well-pruned orchard to \SI{65.2}{\percent} in a dense orchard without changing the detector~\cite{appleharvester}.

Prior blueberry perception studies have focused primarily on maturity classification, detection, and counting~\cite{li2025field, ni2020deep,
sun_bmdnet_2025}, 
with recent work extending to 3D localization~\cite{zhao_blueberry3d_2025}. However, these methods have not been integrated into closed-loop robotic harvesting systems.
For an RGB-D camera, fruit-center localization RMSE grows from \SI{3.13}{\milli\meter} at a capture distance of \SI{30}{\centi\meter} to \SI{24.72}{\milli\meter} at \SI{150}{\centi\meter}~\cite{liu_dualview_2025}, exceeding the alignment tolerance of a berry-scale gripper unless the cluster pose is refined at close range.

\subsection{Integrated Robotic Harvesting Systems}
The design of an integrated robotic harvesting system depends strongly on the characteristics of the target crop.
Integrated autonomous harvesting is established for larger separable fruits, including apples~\cite{appleharvester, li_multi-arm_2023, zhu_advancement_2025}, strawberries~\cite{strawberryharvester}, mango~\cite{mangoharvester}, and tomato~\cite{ansari_integrated_2026}. These systems typically follow a single-fruit picking workflow in which each target is localized, approached, and detached individually, at cycle times of roughly \SIrange{6}{25}{\second} and success rates of \SIrange{70}{90}{\percent} (Table~\ref{tab:systems}). 
Cluster-level harvesting additionally shifts maturity selectivity from perception to the end-effector, which must detach mature berries through controlled contact force. 
To our knowledge, no field-evaluated system integrates cluster-level targeting, selective detachment, and completion detection for fresh-market blueberries harvesting.

%% file: design.tex
\section{Mechatronic Design of the CLASP Platform}
\label{sec:design}
\begin{figure*}[htp]
    \centering
    \includegraphics[width=\textwidth]{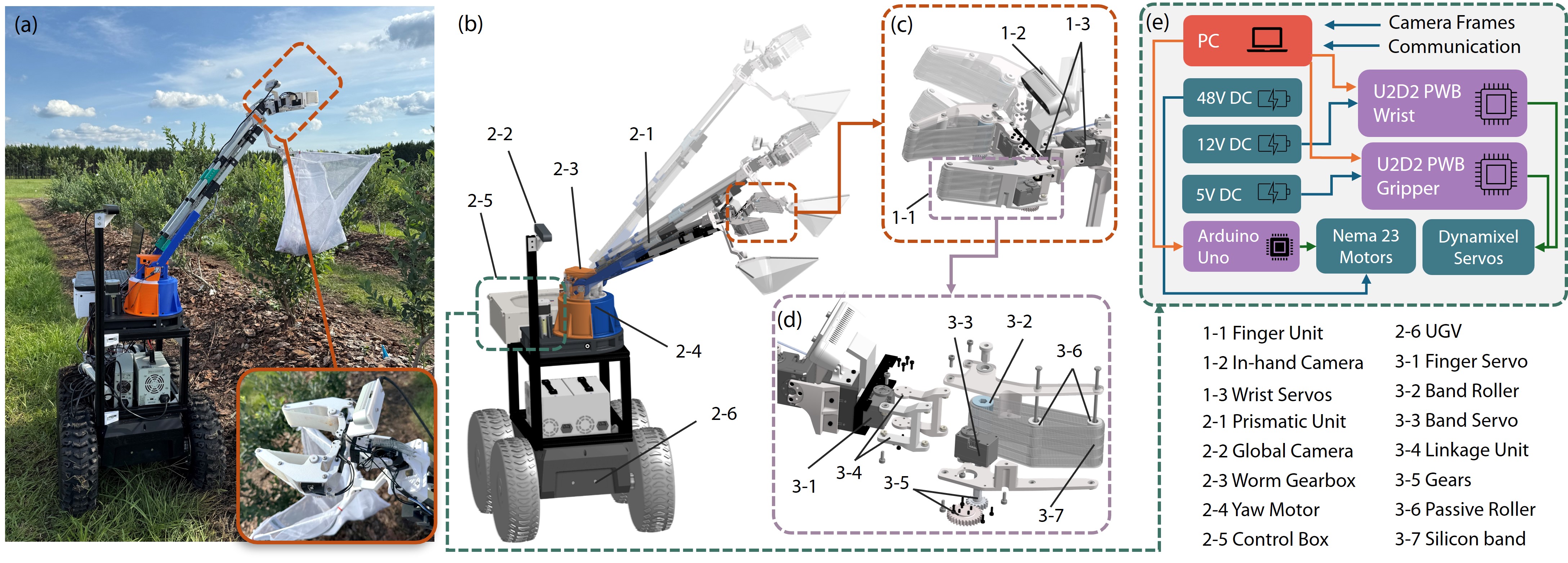}
    \vspace{-5mm}
    \caption{Hardware design of the blueberry harvesting system: (a) system prototype in the field; (b) CAD model of the robotic system; (c) gripper design with three-DOF wrist; (d) exploded view of the finger unit; (e) electronic system.}
    \vspace{-5mm}
    \label{fig:hardware_design}
\end{figure*}
\subsection{Design of the Robotic Manipulator}

The robot employs a six-degree-of-freedom (6-DoF) serial manipulator with an RRPRRR architecture, comprising five revolute joints and a prismatic extension joint at the third position. The layout is shown in Fig.~\ref{fig:hardware_design}(a) and (b), and the Denavit--Hartenberg parameters are listed in Table~\ref{tab:kinematics}.

The manipulator combines high-torque stepper actuation at the proximal joints with compact servo actuation for extension and wrist articulation.
The azimuth joint is driven by a NEMA 23 stepper motor through a 47:1 gearbox and a timing belt. 
The elevation joint is driven by a NEMA 23 stepper motor through a 40:1 self-locking worm gearbox, holding the arm in place without power and preventing uncontrolled motion during power loss.
Downstream of these two joints, a Dynamixel XC430 servo drives the extension joint through a timing belt. The radial reach extends from \SI{937}{\milli\meter} to \SI{1333}{\milli\meter}. The wrist provides roll, pitch, and yaw through one Dynamixel 2XC430 dual-axis servo and one Dynamixel XC430 servo. Joints $\theta_4$, $\theta_5$, and $\theta_6$ rotate about the wrist frame $x$, $y$, and $z$ axes respectively, with $z$ along the approach direction. The complete platform was constructed at a total component cost of approximately only USD~3326.

The manipulator is designed to balance cost, workspace, and orientation dexterity. High-torque proximal actuation provides load-bearing capacity, while compact distal servos reduce moving mass. AS5600 encoders provide position feedback for the stepper-driven joints, enabling closed-loop monitoring with low-cost actuators. This architecture yields a large workspace without long links or high holding torques. It also preserves the orientation flexibility required in dense, occluded canopies.

\begin{table}[!t]
  \caption{Denavit--Hartenberg Parameters and Joint Limits of the 6-DOF Manipulator (Nominal CAD Values)}
  \label{tab:kinematics}
  \centering
  \footnotesize
  \setlength{\tabcolsep}{3pt}
  \begin{tabular}{@{}l c c c c c@{}}
    \toprule
    Joint & $\theta_i$ (deg) & $d_i$ (mm) & $a_i$ (mm) & $\alpha_i$ (deg) & Range (deg) \\
    \midrule
    Azimuth joint       & $\theta_1$          & $232.78$        & $0$      & $-90$ & $[-90, 65]$ \\
    Elevation joint & $\theta_2$          & $0$             & $-36.69$ & $90$  & $[-15, 90]$ \\
    Extension joint   & $90$                & $764.25{+}d_3$  & $0$      & $90$  & $[0, 396]$ \\ 
    Wrist roll     & $\theta_4{-}90$     & $0$             & $-24$    & $90$  & $[-30, 30]$ \\
    Wrist pitch    & $-\theta_5{-}90$    & $0$             & $0$      & $90$  & $[-60, 20]$ \\
    Wrist yaw      & $\theta_6{+}180$    & $148$           & $0$      & $0$   & $[-45, 45]$ \\
    \bottomrule
  \end{tabular}
  \vspace{-5mm}
\end{table}

\subsection{Design of the SARB-Gripper}

\begin{figure}[h]
    \centering
    \includegraphics[width=0.86\linewidth]{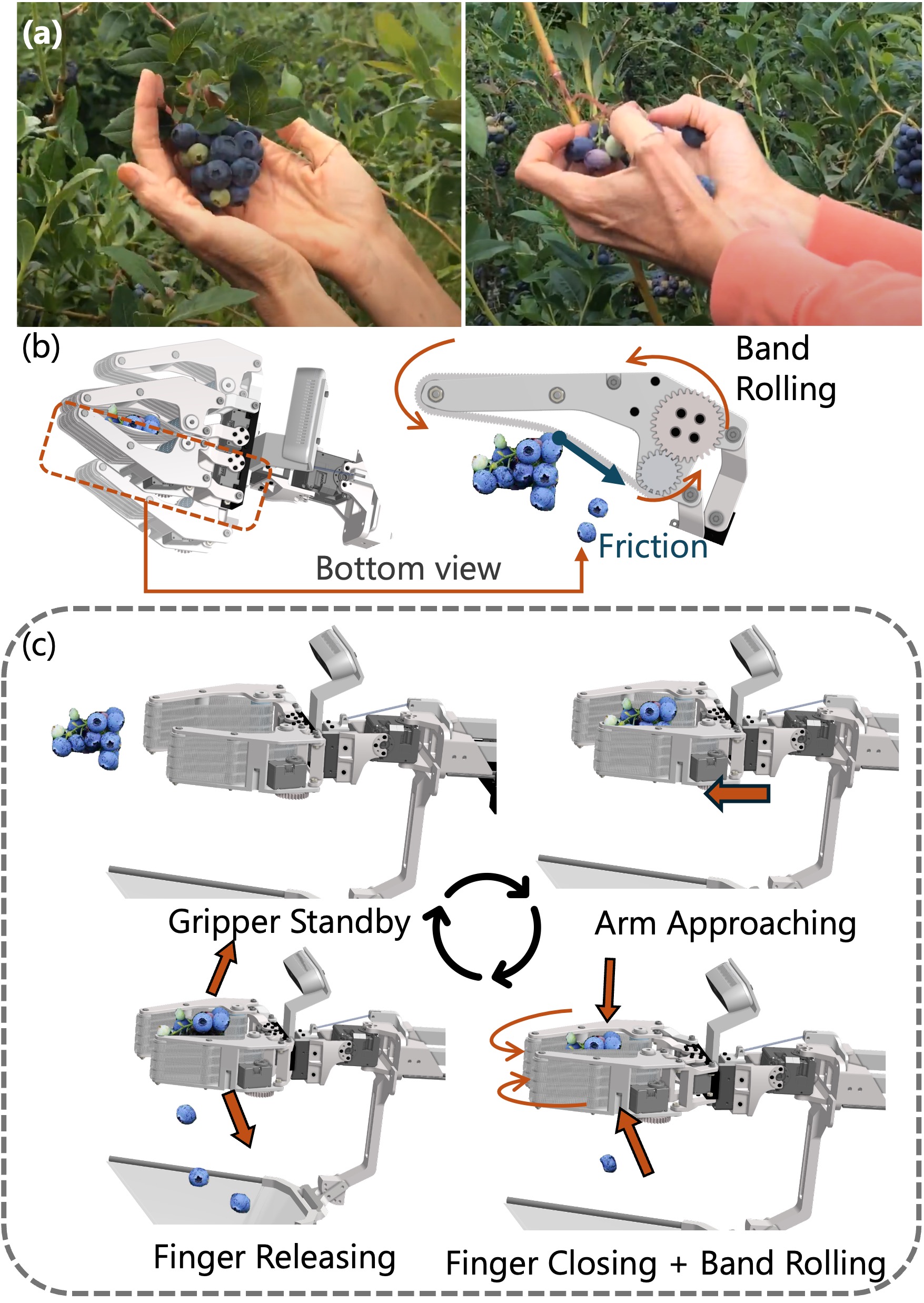}
    \vspace{-1.0 em}
    \caption{Illustration of the grasping workflow, finger-closure movements, and harvesting mechanism: (a) inspiration from a human picker's harvesting demonstration; (b) finger-closure movements and harvesting mechanism; (c) grasping workflow for the gripper design.}
    \label{fig:gripper_illustration}
    \vspace{-1.2 em}
\end{figure}

The Soft Active Rolling-Band Gripper (SARB-Gripper) draws inspiration from the motion used by human pickers. They use the thumb to gently pull or rub berries against the supporting stem to detach them, while the palm is positioned underneath to catch the detached fruit (Fig.~\ref{fig:gripper_illustration}(a)). The thumb contacts several berries at once and does not target them individually. The ripe berries separate under applied load, while the unripe berries remain attached. This picking strategy motivates two coupled functions in the gripper: compliant engagement with the cluster and active tangential motion for fruit detachment.
\begin{figure*}[!h]
    \centering
    \includegraphics[width=\linewidth]{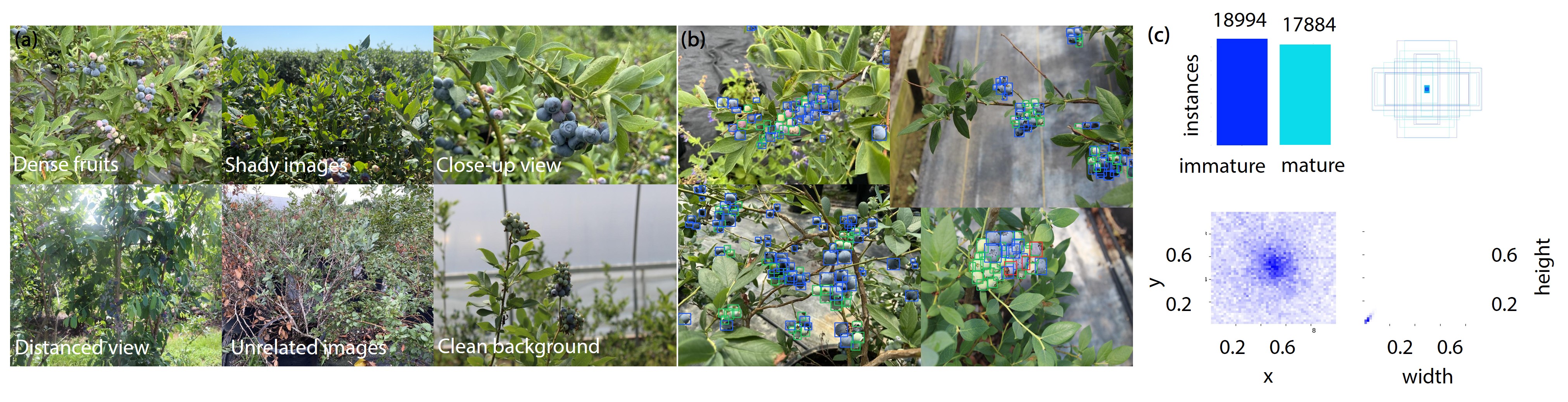}
    \vspace{-2.8 em}
    \caption{Dataset configuration: (a) sample gallery of the dataset; (b) annotation examples from the dataset using bounding boxes; (c) training-set label statistics, showing the per-class instance distribution and the normalized bounding-box dimension distribution.}
    \label{fig:dataset_configuration}
    \vspace{-1.4 em}
\end{figure*}

As shown in Fig.~\ref{fig:hardware_design}(c), the proposed SARB-Gripper consists of two fingers, each equipped with a soft rolling band made of Dragon Skin~20 silicone. Each band is independently actuated by a Dynamixel XC330 servo through a 0.6:1 gear transmission. The two fingers are driven by separate Dynamixel XC330 servos through four-bar linkages (Fig.~\ref{fig:hardware_design}(d)). The gripper provides a maximum finger opening of \SI{100}{\milli\meter}, and a band contact length of \SI{100}{\milli\meter}. During harvesting, the fingers are first closed around the target cluster to establish compliant contact. Actuating the bands drives relative sliding at the fruit--band interface, so interfacial friction detaches the contacted berries at the berry--pedicel junction (Fig.~\ref{fig:gripper_illustration}(b)). 
The rolling bands engage multiple berries within the cluster, enabling continuous detachment without individually locating and picking each fruit.
The detached berries then fall into a soft net stretched and placed below the gripper. An Intel RealSense D435 camera is also mounted above the gripper to provide a close-range eye-in-hand view. The complete operating flow is illustrated in Fig.~\ref{fig:gripper_illustration}(c). 

\subsection{Electronic Architecture}
The electronic architecture is shown in Fig.~\ref{fig:hardware_design}(e). Separate voltage rails provide \SI{48}{\volt} for stepper motors, \SI{12}{\volt} for the Dynamixel XC430 and 2XC430 servos, and \SI{5}{\volt} for the Dynamixel XC330 servos. High-level perception and control run on a laptop equipped with an Intel Core i7-10875H CPU operating at \SI{2.30}{\giga\hertz} and an NVIDIA Quadro T1000 GPU. 
The laptop communicates with the cameras, servo controllers, and microcontroller over USB. The stepper motors are driven by DM556Y drivers using step-and-direction signals from the Arduino, which also reads the AS5600 joint encoders through an I2C multiplexer and the limit switches through digital inputs. 

%% file: perception.tex
\section{Blueberry Detection Dataset and Model}
\label{sec:perception}
Cluster detection provides the visual input for localization and harvesting. It must handle small, dense, and frequently occluded berries under varying field illumination. A multi-season field dataset is therefore developed, and a dedicated berry detector is trained for the harvesting pipeline.

\subsection{Dataset Collection and Configuration}
\label{sec:dataset}
Our dataset was collected over four growing seasons (2023--2026) at three research farms in the U.S. (Poplarville and Starkville, Mississippi; Alapaha, Georgia) and spans four rabbiteye cultivars. Both field-grown and container-based blueberries are included in the dataset, covering research and commercial settings. Images were captured with five cameras of differing focal length and resolution: three smartphone cameras, a Canon EOS~4000D, and an Intel RealSense~D435. 
The dataset therefore spans variations in season, site, cultivar, camera, and time of day.

We annotated an instance-segmentation dataset\footnote{Dataset is available at this
\href{https://app.roboflow.com/clasp-workspace/clasp_training_dataset/1}{link}.} on the Roboflow platform~\cite{roboflow} with three ripeness classes: immature, semi-mature, and mature. The semi-mature class was then merged with the immature class for training because it only represented  \SI{1.9}{\percent} of all annotations and showed ambiguous visual characteristics. 
Small and densely packed berries, including partially occluded instances, were annotated individually to capture the incomplete appearances commonly encountered in the field.
The model was trained and evaluated on a 672-image subset. An additional 40 images were subsequently annotated, resulting in a final open-source dataset of 712 images.
A sample gallery and representative annotations are shown in Fig.~\ref{fig:dataset_configuration}(a) and~(b).

\subsection{Image Data Preprocessing and Augmentation}
We applied offline augmentation to the training data to improve the detection robustness.
Images were auto-oriented to normalize EXIF rotation metadata and resized within \num{1,280}$\times$\num{720}~px while preserving the aspect ratio. The dataset was divided into training, validation, and test splits of \num{940}, \num{134}, and \num{68} images, with only the training set expanded through \(2\times\) offline augmentation. 
To account for variations in sensing conditions throughout field operation, the offline augmentation includes exposure changes of \(\pm\SI{9}{\percent}\), camera-gain variation of \num{0.05}, and additive noise affecting up to \SI{1.09}{\percent} of pixels.
The resulting training set contains \num{18,994} immature and \num{17,884} mature berry instances, with the label statistics shown in Fig.~\ref{fig:dataset_configuration}(c).

\subsection{Detection Model Training Procedure}
\label{sec:training}
Using the offline augmented training dataset described above, we fine-tune a COCO-pretrained YOLO26n (\num{2.4}M parameters)~\cite{jocher2026ultralyticsyolo26unifiedrealtime} with an input resolution of 
\num{1,280}$\times$\num{1,280}~px.
Because each berry occupies only a small fraction of the frame (Fig.~3(c)), the resolution is set above the library default to preserve small-target features, while the compact model keeps on-board inference efficient.

Unlike the offline augmentation that simulates camera and sensing variations, training-time augmentation primarily modifies scene appearance and is configured to preserve color-encoded ripeness cues and the characteristics of densely packed small targets.
Hue jitter is set low (\num{0.015}) because ripeness is encoded by color, whereas saturation and brightness jitter are raised (\num{0.5}) to span the illumination range of a harvesting session. Mosaic and MixUp are disabled for the final \num{15} epochs to allow the model to converge on the natural scale and distribution of densely packed berries.
\begin{figure*}[h]
    \centering
    \includegraphics[width=0.95\linewidth]{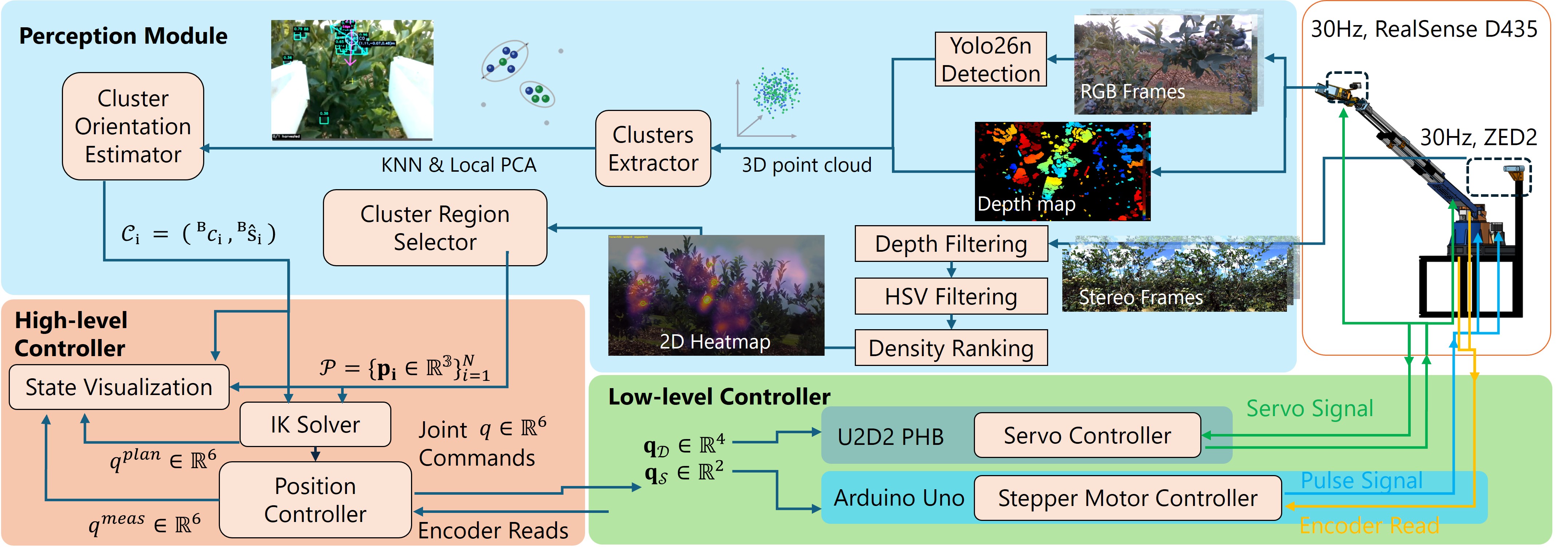}
    \vspace{-0.8 em}
    \caption{Blueberry autonomous harvesting pipeline illustration.}
    \label{fig:autonomous_pipeline}
    \vspace{-0.8 em}
\end{figure*}

Because the harvesting controller requires only berry counts and approximate locations, the instance masks were converted to axis-aligned bounding boxes for training. The remaining training settings follow standard practice. Training ran up for \num{300} epochs with AdamW optimizer with a cosine learning-rate schedule from \(10^{-3}\) to \(10^{-5}\) and early stopping with a patience of \num{50}.
Model checkpoints were selected on the validation split, while the test split was held out for final evaluation.

%% file: control.tex
\section{Autonomous Harvesting Control Pipeline}
\label{sec:control}
In this section, we present the perception, high-level control, and low-level control framework for autonomous cluster-level harvesting, as shown in Fig.~\ref{fig:autonomous_pipeline}.

\subsection{Global-to-Local Perception}
The perception module for cluster localization operates at two spatial scales. A ZED~2 stereo camera first provides coarse global localization, producing a set of candidate cluster 3D regions $\mathcal{P} = \{\mathbf{p}_i \in \mathbb{R}^3\}$ from the RGB and depth streams.

For each candidate region, a scan pose is defined \SI{140}{\milli\meter} from the region along the horizontal direction toward the robot base, with the camera facing that cluster region. Kinematic feasibility is then evaluated using the inverse kinematics solver described below. 
Candidate poses satisfying the position and approach-orientation requirements are added to the scan pose list, while those with infeasible approach orientations are rejected to avoid lateral contact with surrounding branches.

After the arm reaches an accepted scan pose, perception transitions from global localization to close-range cluster scanning using the Intel RealSense D435 on the end-effector. If no cluster is detected over several consecutive frames, the wrist steps through discrete pitch and yaw offsets of up to $40^{\circ}$ until one cluster appears in view. Bounding boxes predicted by the YOLO26n model are then projected through the depth map to obtain the corresponding 3D cluster point clouds.

These point clouds provide the geometry needed to estimate how each cluster is oriented. The orientation is described by a unit vector $\hat{\mathbf{s}} \in \mathbb{R}^{3}$, inferred from the berry distribution because the thin stem is largely occluded. The cluster point cloud is filtered using a local K-Nearest Neighbors (KNN) search, and $\hat{\mathbf{s}}$ is estimated as the leading principal component from Principal Component Analysis (PCA). It is expressed as ${}^{T}\hat{\mathbf{s}}$ in the tool frame $\{T\}$ and ${}^{B}\hat{\mathbf{s}}$ in the base frame $\{B\}$.

The estimated cluster position and orientation are combined into a geometric descriptor for subsequent motion planning. Each scan yields $N$ clusters that are tracked across frames within that scan, with the set reinitialized for each new scan. The $i$-th cluster is represented as
\begin{equation}
    \mathcal{C}_i = ( {}^{B}\mathbf{c}_i,\, {}^{B}\hat{\mathbf{s}}_i ), \qquad i\in\{1,\dots,N\}
\end{equation}
where ${}^{B}\mathbf{c}_i \in \mathbb{R}^{3}$ is the centroid expressed in the base frame $\{B\}$, obtained by transforming the final scan-frame estimate using the camera-to-base transform ${}^{B}\mathbf{T}_{C}(\mathbf{q}_{\mathrm{scan}}) \in SE(3)$. 

\subsection{Grasp Orientation Planning}

\begin{figure}[h]
  \centering
  \includegraphics[width=0.8\linewidth]{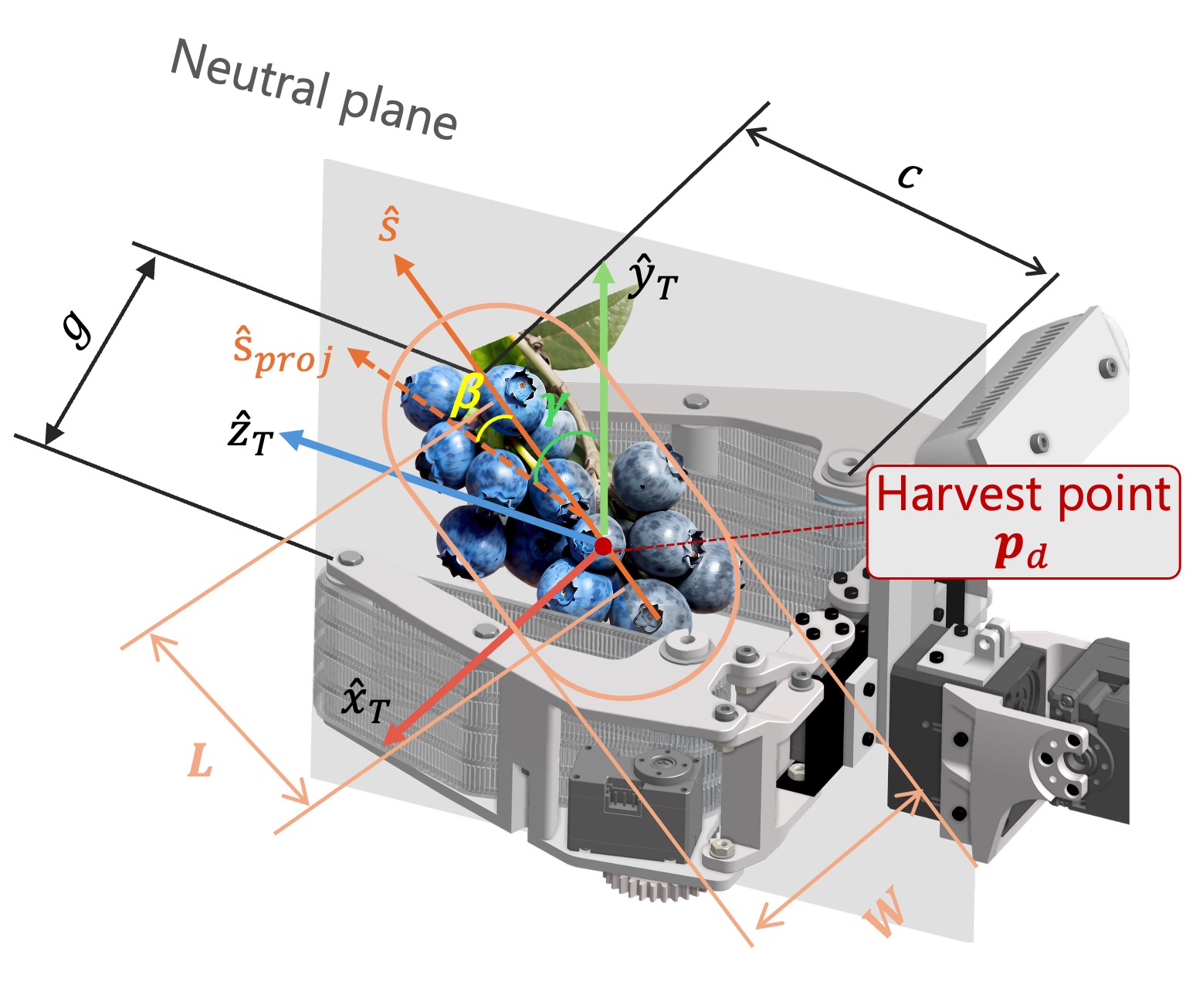}
  \vspace{-5mm}
  \caption{Tool frame $\{T\}$ and cluster geometry at the harvest point $\mathbf{p}_d$. The fingers open by $g$ along the closure axis $\hat{\mathbf{x}}_T$, the gripper approaches along $\hat{\mathbf{z}}_T$, and the transverse axis $\hat{\mathbf{y}}_T$ spans the contact region between the fingers over an effective band contact length $c$. The $\hat{\mathbf{y}}_T$--$\hat{\mathbf{z}}_T$ plane forms the neutral plane. The cluster has stem axis $\hat{\mathbf{s}}$, length $L$, and width $W$. }
  \label{fig:gripper_frame}
  \vspace{-3mm}
\end{figure}

The cluster descriptor specifies the position and orientation of each cluster, but a corresponding gripper orientation must be determined for harvesting. Grasp orientation planning therefore seeks to align the gripper with the estimated cluster axis $\hat{\mathbf{s}}$. The resulting misalignment is decomposed into two components. The out-of-plane misalignment deviation of the cluster axis is $\beta=\angle(\hat{\mathbf{s}},\hat{\mathbf{s}}_{\mathrm{proj}})$ and the in-plane misalignment deviation is $\gamma=\angle(\hat{\mathbf{s}}_{\mathrm{proj}},\hat{\mathbf{y}}_T)$, where $\hat{\mathbf{s}}_{\mathrm{proj}}$ is the projection of $\hat{\mathbf{s}}$ onto the neutral plane (Fig.~\ref{fig:gripper_frame}). 
Both misalignment components vanish when the cluster axis ${}^{T}\mathbf{s}$ aligns with the transverse tool axis $\hat{\mathbf{y}}_T$, which defines the condition that the ideal grasp configuration must satisfy:
\begin{equation}
  {}^{T}\hat{\mathbf{s}} \times \hat{\mathbf{y}}_T=\mathbf{0}.
  \label{eq:orientation-condition}
\end{equation}
In practice, exact alignment isn't always achievable due to errors in $
\hat{\mathbf{s}}$ and the kinematic limits of the arm, so nonzero $\beta$ and $\gamma$ must be tolerated within limits set by three geometric conditions. 
First, the cluster must fit through the finger opening during insertion, requiring its projected width along the closure axis to satisfy
$L\sin\beta + W \leq g$.
Second, the two rolling bands must contact opposite sides of the cluster symmetrically to avoid resulting moments that bend or twist the stem, requiring the cluster axis to lie in the neutral plane at $\beta = 0$. Third, the cluster must remain within the effective rolling region, so its extent along the band-transport direction satisfies $L\sin\gamma + W \le c$. Berries near the edge of the contact region otherwise experience insufficient rolling distance for reliable detachment.

The ideal alignment in \eqref{eq:orientation-condition} maximizes the geometric margins for both insertion and rolling contact. As the cluster length $L$ increases, these margins decrease, making longer clusters less tolerant to orientation errors and motivating cluster-specific orientation adjustment. However, the achievable adjustment is constrained by the fixed approach direction $\hat{\mathbf{z}}_T$. Wrist roll is therefore used to minimize the deviation from the ideal alignment in \eqref{eq:orientation-condition}. Specifically, the roll drives the out-of-plane deviation $\beta$ to zero, while the remaining in-plane deviation $\gamma$ is determined by the approach direction. The resulting orientation provides the closest achievable alignment and determines the wrist roll commanded for each cluster.

\subsection{Joint-Space Arm Control}

The arm controller takes the cluster descriptor and the planned grasp orientation and produces the joint commands that place the gripper on the target. The controlled point is the tool center point (TCP), located \SI{148}{\milli\meter} distal to the wrist pitch axis ($d_6$ in Table~\ref{tab:kinematics}). The tool approach vector $\mathbf{a}(\mathbf{q})$ is defined as the tool-frame $z$-axis, corresponding to the direction along which the gripper engages the cluster.

\subsubsection{Inverse Kinematics (IK)}
The proximal azimuth--elevation--extension chain admits a closed-form position solution, but the full chain does not satisfy Pieper's criterion~\cite{pieper1968}. Because axis~4 is offset by $|a_4|=\SI{24}{\milli\metre}$ from the intersection of axes~5 and~6, the wrist is non-spherical and position and orientation remain coupled. We therefore solve the full-chain IK numerically, initialized by the closed-form shoulder solution~\cite{zhang_innovative_2025}. For a desired harvest pose, we solve the box-constrained nonlinear least-squares problem:
\begin{equation}
\begin{array}{ll}
& \min
\lVert \mathbf{r}(\mathbf{q}) \rVert_2^2 \\
\text { s.t. } & \mathbf{q}_{\min}\le\mathbf{q}\le\mathbf{q}_{\max}
\label{eq:ik_residual}
\end{array}
\end{equation}
where
\begin{equation} 
\mathbf{r}(\mathbf{q}) =
\begin{bmatrix}
\bigl(\mathbf{p}_d - \mathbf{p}(\mathbf{q})\bigr) / \varepsilon_p \\
\boldsymbol{\phi}\!\left(\mathbf{R}_d^\top \mathbf{R}(\mathbf{q})\right) / \varepsilon_R \\
\bigl(\mathbf{u}(\mathbf{q}) - \hat{\mathbf{z}}_T\bigr) / \varepsilon_u \\
\theta_5 / \varepsilon_\theta
\end{bmatrix}.
\end{equation}
The per-joint limits $\mathbf{q}_{\min}$ and $\mathbf{q}_{\max}$ are defined in Table~\ref{tab:kinematics}, with the inequalities taken component-wise. Here, $\mathbf{p}_d$ and $\mathbf{R}_d=[\hat{\mathbf{x}}_T\ \hat{\mathbf{y}}_T\ \hat{\mathbf{z}}_T]$ denote the desired harvest position and tool orientation, respectively. The approaching axis $\hat{\mathbf{z}}_T$ is selected normal to the estimated cluster axis $\hat{\mathbf{s}}$ and directed from the robot toward the cluster, while the roll about $\hat{\mathbf{z}}_T$ is determined from the cluster orientation as described in Section~\ref{sec:control}-A. The operator $\boldsymbol{\phi}(\cdot)$ maps the rotation error to its axis-angle representation. In addition, $\mathbf{u}(\mathbf{q})$ is the unit vector along $\partial\mathbf{p}/\partial d_3$, representing the instantaneous TCP  translation induced by extension of the prismatic joint.

The four residual blocks in $\mathbf{r}(\mathbf{q})$ respectively enforce TCP position, tool orientation, prismatic travel-axis alignment, and a preferred wrist configuration. To reduce collision risk from the lateral sweeping motion of the azimuth joints, the travel-axis term favors configurations in which the final approach is achieved primarily through prismatic extension along the desired approach direction. Aligning $\mathbf{u}(\mathbf{q})$ with $\mathbf{a}_d$ therefore makes the final approach approximately collinear with the tool axis. We then use different scaling parameters for the residual blocks: $\varepsilon_p=\SI{5}{\milli\metre}$, $\varepsilon_R=\ang{10}$, $\varepsilon_u=\ang{5}$. The tighter travel-axis tolerance places greater emphasis on straight prismatic insertion. The final term uses $\varepsilon_\theta=\ang{33}$ as a weak regularizer that favors an extended wrist when multiple configurations otherwise satisfy the task constraints.

To initialize the numerical IK, the target cluster position is used to evaluate the closed-form solution of the proximal three joints and obtain a nominal configuration. Multiple initial guesses are formed from this nominal solution, a fixed set of seeds spanning several elevation and wrist angles, and a warm start from the current wrist. Each of the 12 initial guesses is solved as a box-constrained nonlinear least-squares problem using a trust-region-reflective algorithm (\SI{27.7}{\milli\second} on average), and the converged solution with the smallest TCP position error is selected. Over 500 harvest poses sampled across the reachable workspace, the complete solution required \SI{334\pm129}{\milli\second} on the computer described in Section~\ref{sec:design}. The solver is invoked once per harvest pose rather than within a feedback loop, so this cost is negligible relative to the motion time of a harvest cycle.

\subsubsection{Motion Execution}
For each feasible target pose, the IK solver provides a target joint configuration $\mathbf{q}_d$. Motion is executed through a predefined joint sequence to reduce collisions with the surrounding canopy. Before large arm motions, the extension joint retracts to keep the manipulator compact. The azimuth and elevation joints then move toward the target, followed by the wrist joints, and the extension joint extends last to provide an approximately linear approach to the cluster. Joint trajectories are generated by interpolation in joint space, with trapezoidal velocity profiles applied to the azimuth and elevation joints to reduce structural oscillation.

The detected clusters are processed sequentially. For each cluster, the end-effector first moves to its approach pose and then advances to the harvest pose with the planned wrist orientation. After the harvesting action described in Section~\ref{sec:auto-grasp}, the arm returns to the corresponding scan pose
before proceeding to the next cluster.

\subsection{Autonomous Gripper Control for Cluster-Level Harvesting}
\label{sec:auto-grasp}
Selective cluster harvesting requires the gripper to operate within the
detachment-force window identified in Section~\ref{sec:gripper_force}, applying
sufficient load to detach mature berries while remaining below the higher
detachment force of immature fruit. The gripper therefore regulates the
interaction load through motor-current feedback while the rolling bands
progressively detach mature berries.

Because the finger motors operate in current-based position control, motor
current increases when finger motion is resisted by an enclosed cluster and is
therefore used as a surrogate for interaction force. Before harvesting, the
unloaded current of each finger is calibrated as $\mu_i$. Initial contact is
detected when the window-averaged current exceeds
\begin{equation}
  I_{\mathrm{th},i} = \mu_i + I_{\mathrm{cal}},
  \label{eq:contact-threshold}
\end{equation}
where $I_{\mathrm{cal}}$ provides a margin above unloaded motor current and measurement noise caused by gear friction. The current is averaged over a sliding window, so a brief current spike is not detected as contact.

Each finger is independently regulated through its normalized closure $\alpha_i \in [0,1]$, updated at every control tick as
\begin{equation}
  \alpha_i \leftarrow
  \begin{cases}
    \alpha_i - \Delta, & I_i > I_{\mathrm{th},i},\\
    \alpha_i + \Delta, & I_i \leq I_{\mathrm{th},i},
  \end{cases}
  \label{eq:closure-law}
\end{equation}
where $I_{\mathrm{th},i}$ is the active threshold for the current stage.
Thus, each finger closes while the measured load remains below the threshold and opens slightly when the threshold is exceeded, bounding the interaction load. Independent regulation of the two fingers allows the gripper to conform to irregular and off-center clusters. Harvesting begins after bilateral contact is maintained for a sustained interval.

During harvesting, rolling-band motion continuously changes the cluster geometry and corresponding motor load. The fixed contact threshold is therefore replaced by an adaptive threshold. A tracking baseline $\bar{I}_i$, initialized at the onset of harvesting, is updated as
\begin{equation}
    \begin{aligned}
        \bar{I}_i \leftarrow \lambda \bar{I}_i + (1-\lambda) I_i,
      \qquad I_{\mathrm{th},i} = \bar{I}_i + \delta,
    \end{aligned}
  \label{eq:grip-tracking}
\end{equation}
where $\lambda$ sets the tracking rate and $\delta$ the allowable margin above the baseline. The controller thus responds to load changes relative to a slowly varying baseline rather than to the absolute current. Because \eqref{eq:closure-law} has no dead band, each finger makes small opening and closing adjustments about the threshold while the bands roll continuously, maintaining contact with the changing cluster as mature berries detach.

Harvesting terminates based on either current or vision feedback. The current-based condition is satisfied when both finger currents return near their unloaded values $\mu_i$ for a sustained interval, indicating that the cluster no longer obstructs finger closure. For vision-based termination, the fingers periodically open and the detector evaluates the remaining mature berries. Harvesting terminates when the detected count falls below a preset threshold, with the maximum count over several consecutive frames used to reduce single-frame misses. The controller parameters are summarized in Table~\ref{tab:grip_params}.
\begin{table}[!t]
  \caption{Gripper Controller Parameters}
  \label{tab:grip_params}
  \centering
  \small
  \begin{tabular}{@{}l l l@{}}
    \toprule
    Symbol & Value & Description \\
    \midrule
    $\Delta$ & $0.01$ & Closure increment per tick \\
    $I_{\mathrm{cal}}$ & \SI{15}{\milli\ampere} & Approach-threshold margin \\
    $\lambda$ & $0.97$ & Baseline tracking factor (\SI{0.67}{\second}) \\
    $\delta$ & \SI{35}{\milli\ampere} & Harvesting-threshold offset \\
    \midrule
    \multicolumn{3}{@{}l}{\itshape Timing and termination}\\
    Control rate & \SI{50}{\hertz} & Closure update rate \\
    Contact confirmation & \SI{0.3}{\second} & Both fingers, to enter harvest \\
    Release margin & \SI{8}{\milli\ampere} & Above $\mu_i$, for \SI{0.8}{\second} \\
    \bottomrule
  \end{tabular}
\vspace{-5mm}
\end{table}

%% file: analysis.tex
\section{Workspace and Dexterity Analysis}

The autonomous harvesting pipeline requires the manipulator not only to reach a target cluster but also to achieve a suitable gripper orientation defined in \eqref{eq:orientation-condition}. We therefore analyze both the reachable workspace and the orientation dexterity of the robot. The harvest point is the location at which the gripper contacts the cluster, and the tool frame is attached to it, as shown in Fig.~\ref{fig:gripper_frame}. 

At a given harvest position \(P\), multiple joint configurations reach the same harvest point while producing different gripper orientations. We define \(\mathcal{Q}(P)\) as the set of all configurations that reach \(P\) that satisfy the joint limits:
\begin{equation}
\mathcal{Q}(P)=\{\,\mathbf q : \mathbf q_{\min}\le\mathbf q\le\mathbf q_{\max},\; \mathit{FK}(\mathbf q)=P \,\} .
\label{eq:qset}
\end{equation}
where \(\mathit{FK}(\cdot)\) 
is the forward kinematics and \(\mathbf q_{\min}\), \(\mathbf q_{\max}\) are the joint limits defined in Table~\ref{tab:kinematics}.
Each configuration provides a different orientation of the transverse tool axis \(\hat{\mathbf y}_T\). 
We therefore define the achievable orientation set \(\mathcal S(P)\) as all cluster-axis orientations that can be aligned with \(\hat{\mathbf y}_T\) by at least one configuration in \(\mathcal Q(P)\):
\begin{equation}
\mathcal S(P)=
\left\{
\hat{{\mathbf s}}\in\mathbb{RP}^{2}:
\exists\,\mathbf q\in\mathcal Q(P),\;
\hat{{\mathbf s}}\times\hat{{\mathbf y}}_T(\mathbf q)=\mathbf 0
\right\}.
\label{eq:membership}
\end{equation}
Because the cluster axis is unoriented, \(\mathbf s\) and \(-\mathbf s\) are equivalent, giving a total orientation-space measure of \(2\pi\) steradians. The orientation dexterity \(\rho(P)\) is defined as the fraction of all cluster-axis directions contained in \(\mathcal{S}(P)\),
\begin{equation}
\rho(P)=\frac{1}{2\pi}\,\bigl|\mathcal{S}(P)\bigr| \in[0,1] ,
\label{eq:rho}
\end{equation}

The achievable orientation set is evaluated directly from forward kinematics. We uniformly sample \(4\times10^{3}\) cluster-axis directions over \(\mathbb{RP}^2\) and estimate \(\rho(P)\) as the fraction of directions that are achievable at $P$. The metric is evaluated on a $25$-mm grid throughout the reachable workspace. At this grid resolution, the numerical uncertainty in \(\rho\) is bounded by \(\pm1.3\) percentage points at the median and \(\pm2.8\) percentage points in the worst case.

\begin{figure}[!ht]
  \centering
  \includegraphics[width=\linewidth]{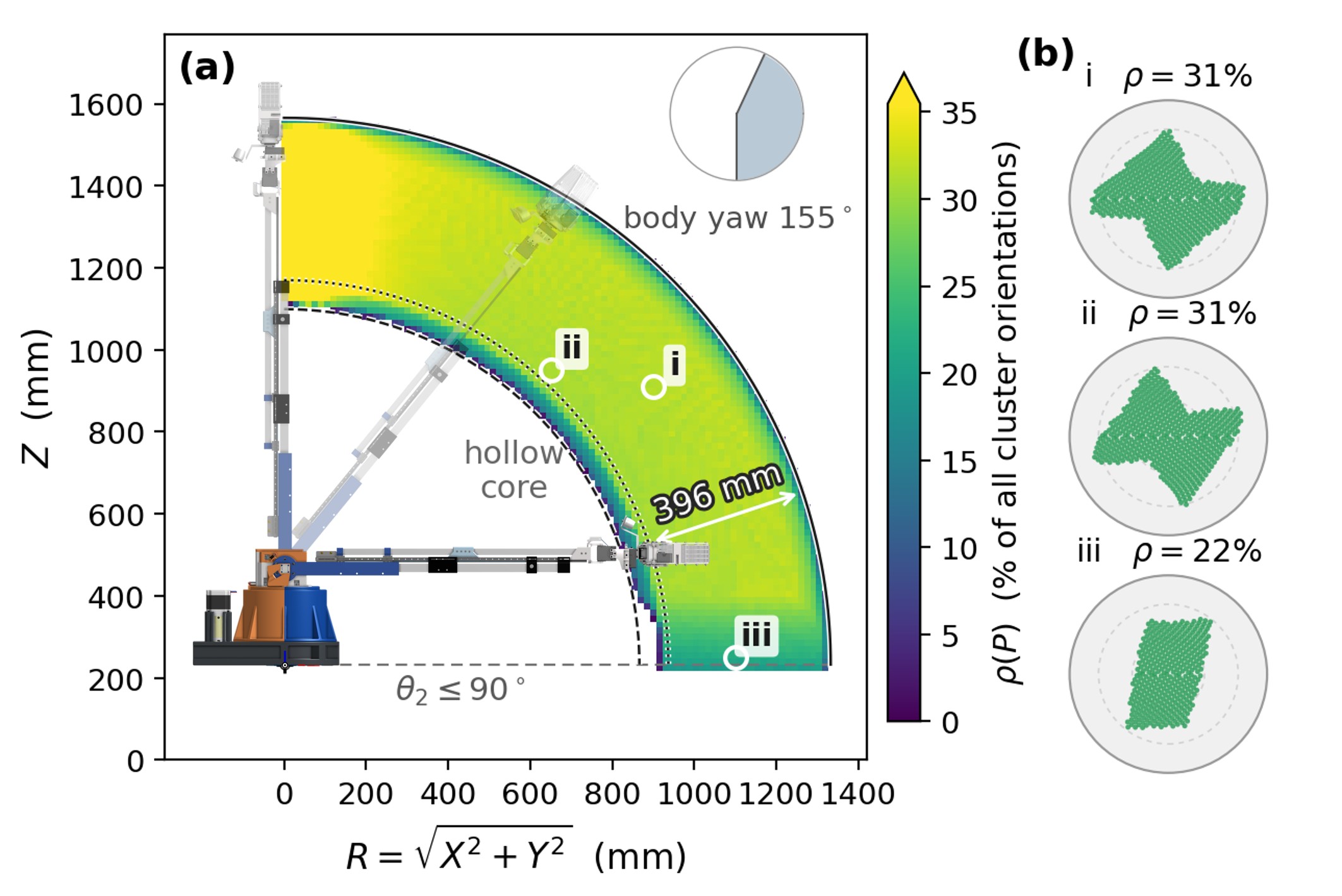}
  \caption{
  Workspace and grasp-orientation dexterity of the harvest point on the $(R,Z)$ section. (a) Reachable envelope colored by the orientation dexterity $\rho(P)$ from (10). (b) Achievable orientation set $\mathcal{S}(P)$ at the three marked positions, shown on a Lambert equal-area chart of $\mathbb{RP}^2$ centered on the gripper axis at the neutral wrist. Parallel directions lie at the center and perpendicular ones at the edge, so the green share of each disc equals $\rho(P)$.
  }
  \label{fig:workspace}
  \vspace{-5mm}
\end{figure}

As shown in Fig.~\ref{fig:workspace}, the orientation dexterity remains largely uniform throughout the reachable workspace. The volume-weighted median is \(31.2\%\) with an interquartile range of \(27.3\%\) to \(31.8\%\), and \(89\%\) of the reachable volume has \(\rho>20\%\). Lower dexterity is concentrated near two workspace boundaries, where joint limits reduce the range of configurations available at a given position. Near the inner radial boundary, prismatic retraction limits elevation joint variation, yielding an average dexterity of \(25.5\%\) between \(866\) and \(1000\)~mm. Below \(400\)~mm, the elevation joint limit reduces the average dexterity to \(23.1\%\). 

%% file: experimentsandresults.tex
\section{Experiments and Results}
\label{sec:results}

\subsection{Perception Test Results}
\label{sec:detection-results}
\begin{figure}
    \centering
    \includegraphics[width=\linewidth]{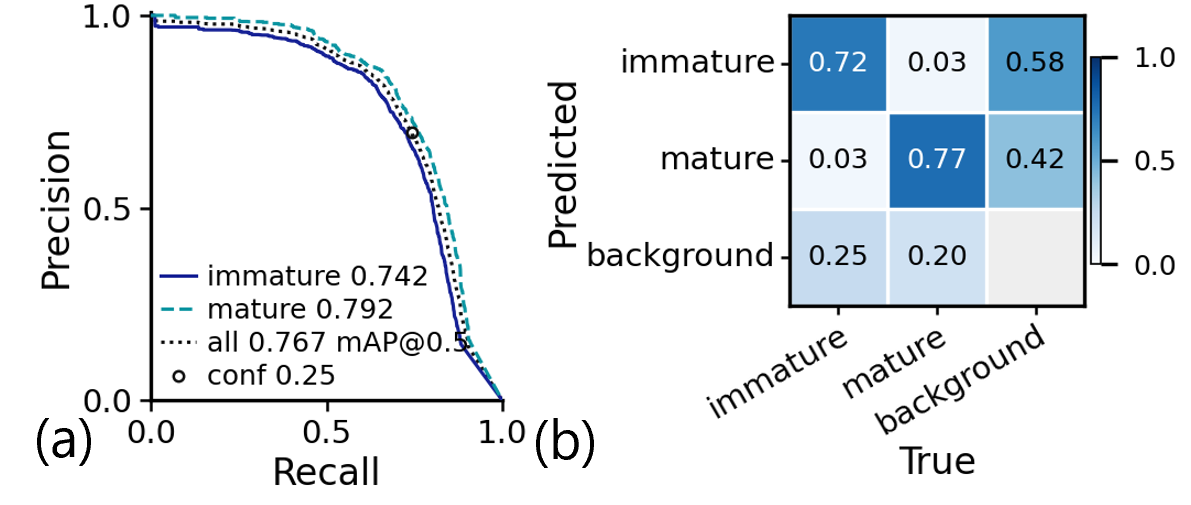}
    \caption{Detection model evaluation on the held-out test split: (a) precision--recall curves at an IoU threshold of $0.5$ for the immature and mature classes and their average; (b) confusion matrix, normalized so each ground-truth column sums to one; columns are ground-truth classes, and rows are predictions.}
    \label{fig:dataset_evaluation}
    \vspace{-6mm}
\end{figure}

\begin{table}[h]
  \caption{Detection Performance on the Held-Out Test Split}
  \label{tab:detection}
  \centering
  \small
  \begin{tabular}{@{}l
                  S[table-format=4.0]
                  S[table-format=1.3]
                  S[table-format=1.3]
                  S[table-format=1.3]
                  S[table-format=1.3]
                  S[table-format=1.3]@{}}
    \toprule
    Class & {Inst.} & {$P$} & {$R$} & {$F_1$} &
    {mAP$_{50}$} & {mAP$_{50\text{--}95}$} \\
    \midrule
    Immature & 1665 & 0.799 & 0.649 & 0.716 & 0.742 & 0.478 \\
    Mature   & 1134 & 0.795 & 0.699 & 0.744 & 0.792 & 0.574 \\
    \midrule
    All      & 2799 & 0.797 & 0.674 & 0.730 & 0.767 & 0.526 \\
    \bottomrule
  \end{tabular}

  \vspace{2pt}
  
  \footnotesize
  Evaluated at \num{1280}~px without test-time augmentation.
  \vspace{-2mm}
\end{table}
As mentioned in Section.~\ref{sec:perception}, the detection model needs to provide accurate and robust berry detection under field conditions. 
We therefore evaluated the model on the test split, which contains $68$ images with $2{,}799$ annotated instances, with an average of $41$ berries per image. Our detection model reaches an mAP$_{50}$ of $0.767$ and a stricter mAP$_{50\text{--}95}$ of $0.526$, averaged across both classes. 
At the operating point that maximizes the \(F_1\) score, a metric that balances precision and recall, the model achieves a precision of \(0.797\) and a recall of \(0.674\) Table~\ref{tab:detection}).
The corresponding precision--recall curves are shown in Fig.~\ref{fig:dataset_evaluation}(a). 
Among these metrics, mAP$_{50}$ is particularly relevant to harvesting because the controller relies more on correctly detecting and counting berries than on highly precise bounding-box localization.

The mature class achieves higher recall and mAP than the immature class, while maintaining comparable precision (Table~\ref{tab:detection}). 
This higher detection performance for mature fruit likely results from their more distinctive and consistent appearance, whereas the immature class also includes partially ripened berries with greater visual variation.
This is particularly important for harvesting because both target selection and the decision to terminate harvesting a cluster depend on the count of mature berries. 

The normalized confusion matrix of Fig.~\ref{fig:dataset_evaluation}(b) 
distinguishes between two types of detection errors.
Misclassification of maturity is rare, with \SI{3}{\percent} of immature berries predicted as mature and \SI{3}{\percent} of mature berries predicted as immature. The dominant error is the missed detections, with \SI{20}{\percent} mature berries and \SI{25}{\percent} immature berries assigned to the background.

Missed detections matter more than misclassification for harvesting. A single observation may under-count the remaining mature berries and terminate harvesting prematurely. The controller therefore aggregates detections over multiple frames during the inspection window (Section~\ref{sec:auto-grasp}), using the maximum observed count over the window rather than a single frame. Berries that remain occluded throughout the inspection window, however, cannot be recovered by this strategy.

\subsection{Gripper Force Test}
\label{sec:gripper_force}

\subsubsection{Experimental Setup}

Selective harvesting requires a pulling force that detaches mature berries while remaining below the detachment force of immature fruit. Both the gripper pulling force and the berry detachment force were therefore characterized, using the setups in Fig.~8(a) and (b). Each experiment used an ATI Nano17 six-axis force sensor sampled at \SI{100}{\hertz}, with $F_z$ aligned with the harvesting direction.


The pulling force transmitted by the gripper was characterized indoors using the control algorithm described in Section~\ref{sec:control}-D. A spherical object was coupled directly to the force sensor, and three diameters were tested in turn, \SIlist[list-units = single, list-final-separator = {, and }]{23;30;35}{\milli\meter}. For each diameter, we varied the servo current offset $I$ from \SIrange{5}{50}{\milli\ampere} with \SI{5}{\milli\ampere} per increment. With two to four repetitions per setpoint, the variation comprises $79$ trials, with \SIrange{25}{45}{\second} of force data recorded per trial. 
The reported pulling force was obtained as the peak of a \SI{50}{\milli\second} moving average of \(F_z\), suppressing single-sample transients while preserving the physical force peak.
Because the two fingers contact the whole cluster on opposite sides, the measured force was divided by two to estimate the equivalent single-side-contact force applied to individual objects. 

The detachment force was further measured in the field across maturity stages. The sensor was connected to a berry clipper with a soft contact surface. After clipping a berry, the sensor was drawn along the harvesting axis until the berry separated from its stem, and the peak of $F_z$ was recorded as the detachment force.


\subsubsection{Experimental Results}

Selective pick requires that the pulling force be set finely and predictably through the servo current offset $I$ and that, for a given setpoint, the delivered force be repeatable.

\begin{figure}[!t]
  \centering
  \includegraphics[width=0.46\textwidth]{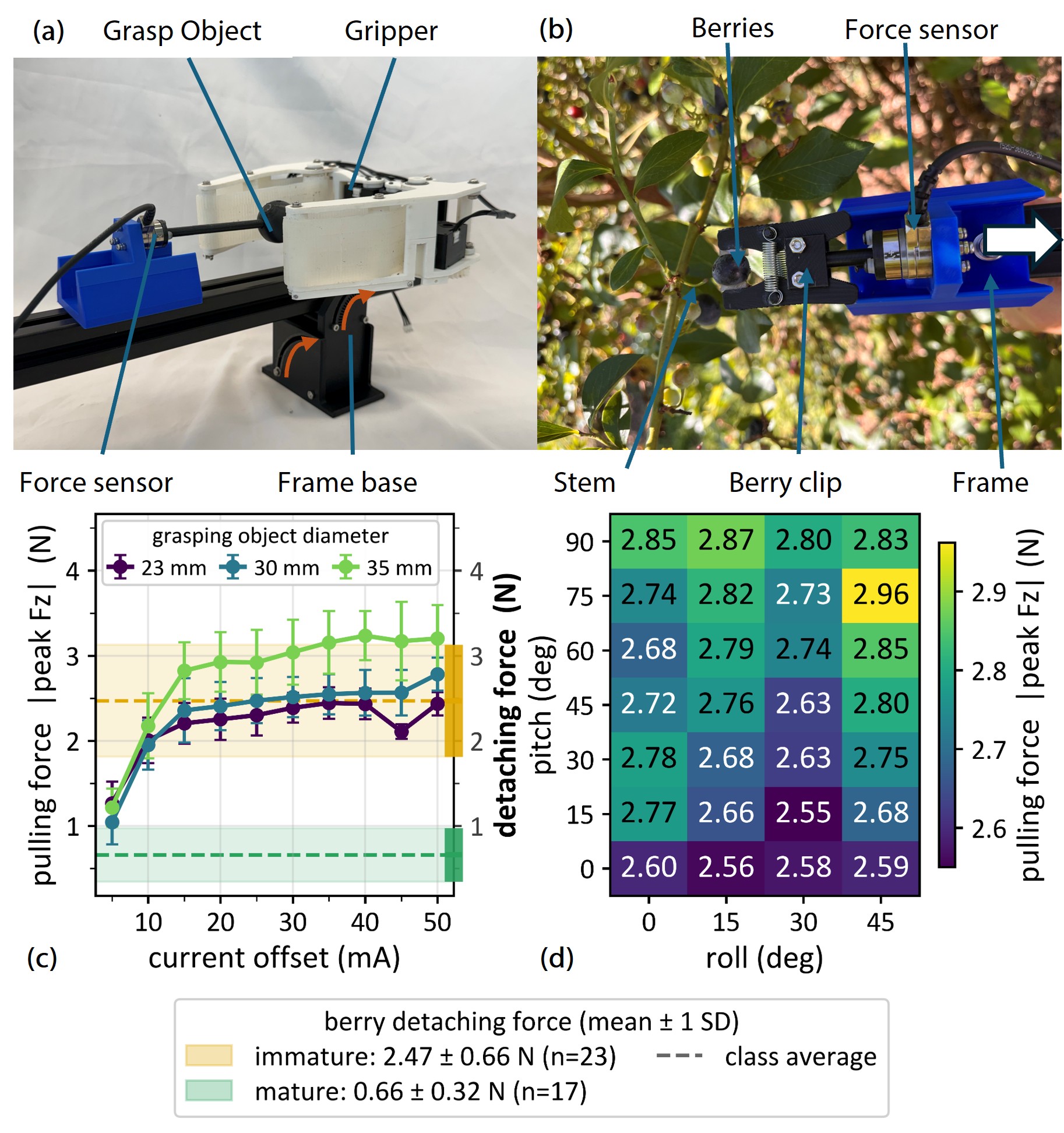}
  \caption{Experiment setup and gripper force characterization: (a) gripper pulling-force measurement and (b) blueberry detaching-force measurement; (c) pulling force ($|\text{peak } F_z|$) versus servo current offset for three grasping-object diameters (left axis), with shaded bands marking the measured berry detaching force (right axis, mean $\pm$ 1 SD) for immature and mature fruit; (d) pulling force over wrist pitch and roll at a fixed \SI{30}{\milli\meter} object and \SI{30}{\milli\ampere} offset.}
  \label{fig:detach_force_summary}
  \vspace{-5mm}
\end{figure}
{As shown in Fig.~\ref{fig:detach_force_summary}(c) the pulling force increases smoothly and monotonically with current, with slopes of \SIrange{0.09}{0.16}{\newton\per\milli\ampere} across the three object diameters. The force also increases with object diameter across the tested range of \SIrange{23}{35}{\milli\meter}.
These trends enable the pulling force to be regulated through the commanded current according to the grasped object size. At higher currents, the force gradually saturates beyond approximately \SI{15}{\milli\ampere}, bounding the load transmitted by the gripper and thereby limiting over-gripping.

At a fixed setpoint, the pulling force varied only \SIrange{1.4}{3.7}{\percent} on average across repeated grasps, confirming repeatability of the commanded force. 
The force control also remained robust to wrist pose, with the pulling force varying by only \SI{0.41}{\newton} (about \SI{15}{\percent} of the mean) across the full pitch and roll grid in Fig.~\ref{fig:detach_force_summary}(d). Together, these results indicate that the gripper can reproducibly regulate the pulling force without requiring precise end-effector orientation.

The horizontal dashed lines and shaded bands in Fig.~\ref{fig:detach_force_summary}(c) show the actual in-field detachment-force measurements for different maturity. Mature berries detached at \SI{0.66\pm0.32}{\newton} and immature berries at \SI{2.47\pm0.66}{\newton}, showing the detaching force difference with no overlap at one standard deviation. The intervening \SI{0.83} {\newton} window allows the pulling force to detach ripe fruit while leaving unripe fruit attached. The achievable force range also depends on object size. The saturation force increases with object diameter, from \SI{2.34}{\newton} at \SI{23}{\milli\meter} to \SI{3.09}{\newton} at \SI{35}{\milli\meter}. The current setpoint can therefore be adjusted according to the measured cluster dimension.
\subsection{In-Field Test Results}

\begin{figure*}
    \centering
    \includegraphics[width=0.9\linewidth]{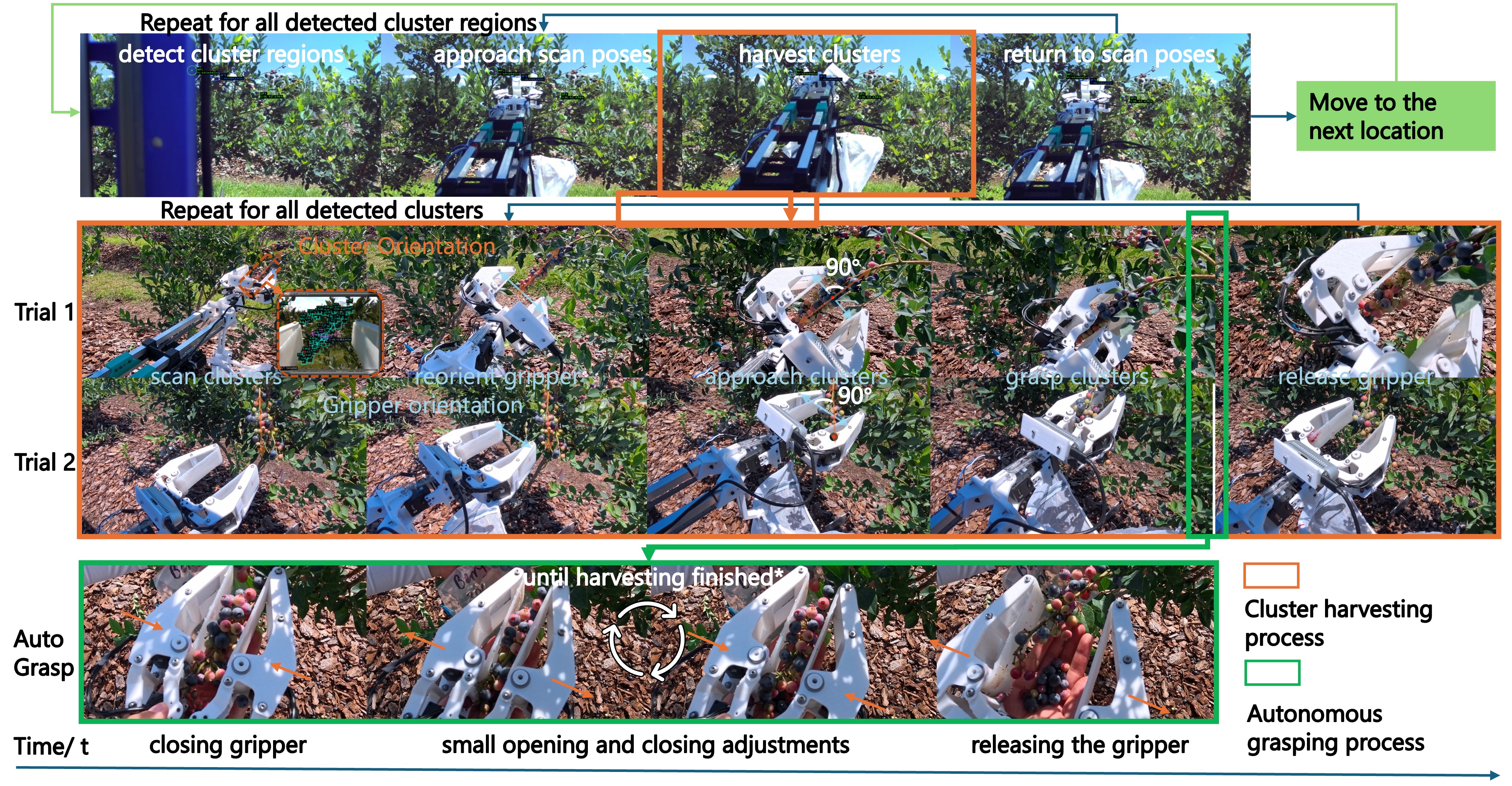}
    \caption{Field demonstration of the autonomous harvesting workflow, read left to right as a time sequence across three nested levels. The platform services each detected cluster regions (top), the arm navigates to scan pose and detect clusters before approaching and grasping it (middle, orange), and the gripper repeats closing and opening cycles until the termination criterion is met (bottom, green).}
    \label{fig:field_exp}
    \vspace{-5mm}
\end{figure*}
System performance was evaluated through field experiments characterizing the crop environment, harvesting performance, post-harvest berry firmness, and gripper throughput.
\subsubsection{Field Test Setup}
Field tests were conducted at the University of Georgia Alapaha Blueberry Research Farm (Alapaha, GA) on three cultivars spanning both major blueberry types grown in the southeastern U.S.: `Alapaha' (southern highbush, \textit{Vaccinium corymbosum} interspecific hybrid) and `Brightwell' and `Vernon' (rabbiteye, \textit{V.~virgatum}).
The robot was mounted on a mobile platform (Rover Robotics MAX) for teleoperated in-field transport.
Human hand picking was also performed as a baseline for comparison. Harvested berries from both robotic and manual picking were subsequently evaluated using a FruitFirm~1000 (CVM, Inc.) to compare post-harvest firmness. Robotically harvested berries were evaluated approximately 5 h after harvesting, whereas manually picked berries were measured immediately.
In addition, 72 clusters were analyzed to characterize the canopy architecture in terms of canopy occlusion and cluster density. The autonomous harvesting performance was evaluated over 25 attempted clusters from 10 rabbiteye bushes.
\subsubsection{Canopy architecture characteristics}
The field environment characterization quantifies the visual challenges faced by the perception pipeline.
Only \SI{40.3}{\percent} (29 of 72) of the clusters surveyed lay within the field of view of the global camera, with the remainder severely occluded by leaves and stems. 
Therefore, this limited observability from a fixed viewpoint demonstrates the need for multi-camera perception strategy.

\subsubsection{Harvesting Success, Efficiency, and Fruit Quality}
\label{harvestingsec}
The harvesting trial measures both end-to-end and modular performance of the full pipeline. Of $25$ harvesting attempts, $23$ succeeded and $2$ failed, giving a success rate of \SI{92}{\percent}. In both failures, the arm could not bring the gripper onto the cluster, so the fruit was unable to be engaged. Performance is therefore limited by perception and approaching accuracy rather than by the harvesting mechanism, which is consistent with the occlusion rate reported above. With $25$ attempts, the result demonstrates the feasibility of the integrated system but should not be interpreted as a comprehensive reliability estimate. Two representative grasping trials are shown in Fig.~\ref{fig:field_exp}.
 
Gripper selectivity and efficiency for cluster-level harvesting were evaluated independently by manually aligning the gripper with each cluster before harvesting. 
The gripper harvested 32 berries per minute, with 90\% of the harvested berries mature, compared with 55 berries per minute and 98\% maturity for a human picker.
This isolates the gripper from the autonomous pipeline and shows that the gripper itself achieves \SI{58}{\percent} of the manual harvesting rate.


As another important aspect of harvest performance, we measured the firmness of individual berries after field tests. Berry firmness provides an indicator of post-harvest fruit quality, as mechanical damage and bruising generally reduce tissue firmness. 
Fruit harvested by the end-to-end fully autonomous system had a mean firmness of \SI{154.21}{\gram\per\milli\meter}, and fruit harvested with autonomous grasping alone had \SI{166.22}{\gram\per\milli\meter}, compared with \SI{195.59}{\gram\per\milli\meter} for the hand-picked reference. This corresponds to a \SI{15}{\percent} reduction for autonomous grasping and a \SI{21}{\percent} reduction for the fully autonomous system. Two effects contribute to this gap. The first is mechanical loading during grasping. The second is the longer interval between autonomous harvesting and later firmness measurement, whereas hand-picked berries were measured immediately,

Because both factors can reduce measured firmness, the observed difference represents an upper bound on the reduction attributable to the harvesting mechanism itself. Section~\ref{sec:discussion} examines the two effects separately.  Even under this conservative interpretation, the observed reduction compares favorably with the \SIrange{16}{26}{\percent} severe bruising rates reported for mechanical harvesters in Section~\ref{sec:introduction}.

%% file: discussion.tex
\section{Conclusions and Discussion}
\label{sec:discussion}
In this paper, we presented CLASP, an autonomous robotic system for selective cluster-level blueberry harvesting through force-regulated interaction. We use ripeness-dependent detachment-force characterization to establish the mechanical basis for selective harvesting, while the soft rolling-band gripper enabled force regulation without dedicated force sensing or per-berry pose estimation. The integrated system achieved a \SI{92}{\percent} autonomous cluster grasping
success rate 
in field trials. In addition, the accompanying multi-season dataset supports detector development across diverse canopy and lighting conditions.

Several opportunities remain for further improving and validating the system. First, autonomous throughput can be increased by optimizing arm trajectories and reducing transit time between harvesting poses. Under manually aligned grasp poses,
the gripper already achieved \SI{58}{\percent} of the manual picking rate, indicating that system-level motion rather than the detachment mechanism itself is a major contributor to cycle time. Second, the current field study evaluated 25 clusters at a single site, and future multi-site experiments will further characterize performance across cultivars, canopy structures, and environmental conditions. Third, autonomously harvested fruit showed a \SI{15}{\percent} to \SI{21}{\percent} reduction in firmness relative to hand-picked fruit. Because the autonomous samples experienced a longer interval before measurement, future time-matched experiments will more precisely distinguish harvesting-induced effects from post-harvest storage and thermal exposure.
